\documentclass{article}

\usepackage[preprint]{neurips_2025}

\newcommand{\x}{\mathbf{x}}

\newcommand{\y}{\mathbf{y}}

\newcommand{\data}{\mathcal{D}}

\newcommand{\reward}{r}

\newcommand{\states}{\mathcal{S}}
\newcommand{\actions}{\mathcal{A}}

\newcommand{\bs}{\mathbf{s}}
\newcommand{\ba}{\mathbf{a}}

\usepackage{times}
 \usepackage[printonlyused]{acronym}
\usepackage[export]{adjustbox} %
\usepackage{algorithm}
\usepackage{algorithmicx}
\usepackage{algpseudocode}
\algrenewcommand\algorithmicindent{0.8em}%
\usepackage{amssymb}
\usepackage{amsmath}
\usepackage{array}
\usepackage{bbm}
\usepackage{booktabs}
\usepackage{caption}
\usepackage{float}
\usepackage{graphicx}
\usepackage[pagebackref=true]{hyperref}
\usepackage{cleveref}
\usepackage{lipsum}
\usepackage{microtype}
\usepackage{multicol}
\usepackage{multirow}
\usepackage{nicefrac}
\usepackage{relsize}
\usepackage{siunitx}
\usepackage{subcaption}
\usepackage{todonotes}
\usepackage{wrapfig}
\usepackage{tikz}
\usepackage{verbatim}
\usepackage{xspace}

\usepackage{amsthm}
\algnewcommand{\IfThenElse}[3]{%
  \State \algorithmicif\ #1\ \algorithmicthen\ #2\ \algorithmicelse\ #3}
 \algnewcommand{\IfThen}[2]{%
  \State \algorithmicif\ #1\ \algorithmicthen\ #2}

\usepackage[utf8]{inputenc} %
\usepackage[T1]{fontenc}    %
\usepackage{url}            %
\usepackage{booktabs}       %
\usepackage{amsfonts}       %
\usepackage{nicefrac}       %
\usepackage{microtype}      %

\definecolor{mydarkblue}{rgb}{0,0.08,0.45}
\hypersetup{ %
    pdftitle={},
    pdfauthor={},
    pdfsubject={},
    pdfkeywords={},
    pdfborder=0 0 0,
    pdfpagemode=UseNone,
    colorlinks=true,
    linkcolor=mydarkblue,
    citecolor=mydarkblue,
    filecolor=mydarkblue,
    urlcolor=mydarkblue,
    pdfview=FitH}

\algnewcommand{\algorithmicforeach}{\textbf{for each}}
\algdef{SE}[FOR]{ForEach}{EndForEach}[1]
  {\algorithmicforeach\ #1\ \algorithmicdo}%
  {\algorithmicend\ \algorithmicforeach}%
\newcommand{\pushright}[1]{\ifmeasuring@#1\else\omit\hfill$\displaystyle#1$\fi\ignorespaces}

\usetikzlibrary{tikzmark}
\usetikzlibrary{calc}
\newcommand{\ALGtikzmarkcolor}{black}%
\newcommand{\ALGtikzmarkextraindent}{2pt}%
\newcommand{\ALGtikzmarkverticaloffsetstart}{-.5ex}%
\newcommand{\ALGtikzmarkverticaloffsetend}{-.5ex}%
\makeatletter
\newcounter{ALG@tikzmark@tempcnta}

\newcommand\ALG@tikzmark@start{%
    \global\let\ALG@tikzmark@last\ALG@tikzmark@starttext%
    \expandafter\edef\csname ALG@tikzmark@\theALG@nested\endcsname{\theALG@tikzmark@tempcnta}%
    \tikzmark{ALG@tikzmark@start@\csname ALG@tikzmark@\theALG@nested\endcsname}%
    \addtocounter{ALG@tikzmark@tempcnta}{1}%
}

\def\ALG@tikzmark@starttext{start}
\newcommand\ALG@tikzmark@end{%
    \ifx\ALG@tikzmark@last\ALG@tikzmark@starttext
    \else
        \tikzmark{ALG@tikzmark@end@\csname ALG@tikzmark@\theALG@nested\endcsname}%
        \tikz[overlay,remember picture] \draw[\ALGtikzmarkcolor] let \p{S}=($(pic cs:ALG@tikzmark@start@\csname ALG@tikzmark@\theALG@nested\endcsname)+(\ALGtikzmarkextraindent,\ALGtikzmarkverticaloffsetstart)$), \p{E}=($(pic cs:ALG@tikzmark@end@\csname ALG@tikzmark@\theALG@nested\endcsname)+(\ALGtikzmarkextraindent,\ALGtikzmarkverticaloffsetend)$) in (\x{S},\y{S})--(\x{S},\y{E});%
    \fi
    \gdef\ALG@tikzmark@last{end}%
}

\apptocmd{\ALG@beginblock}{\ALG@tikzmark@start}{}{\errmessage{failed to patch}}
\pretocmd{\ALG@endblock}{\ALG@tikzmark@end}{}{\errmessage{failed to patch}}

\newtheorem*{theorem-nonlabeled}{Theorem}

\newcommand{\expectation}{\mathbb{E}}

\DeclareMathOperator*{\argmax}{arg\,max \:}

\newcommand{\methodName}{practical sub-optimality\xspace}
\newcommand{\expert}{$V^{ \pi^{*} }(s_0)$\xspace}
\newcommand{\topBest}{$V^{\hat{\pi}^{*}}(s_0)$\xspace}
\newcommand{\topBestFive}{$V^{\hat{\pi}^{*}_{D_{\infty}}}(s_0)$\xspace}
\newcommand{\topBestLocal}{$V^{\hat{\pi}^{*}_{D}}(s_0)$\xspace}
\newcommand{\policyAverage}{$V^{\hat{\pi}^{\theta}}(s_0)$\xspace}

\newcommand{\discountFactor}{\gamma}

\acrodef{AGI}{artificial general intelligence}
\acrodef{ANOVA}[ANOVA]{Analysis of Variance\acroextra{, a set of
  statistical techniques to identify sources of variability between groups}}
\acrodef{ANN}{artificial neural network}
\acrodef{API}{application programming interface}
\acrodef{CACLA}{continuous actor critic learning automaton}
\acrodef{cGAN}{conditional generative adversarial network}
\acrodef{CMA}{covariance matrix adaptation}
\acrodef{COM}{centre of mass}
\acrodef{CTAN}{\acroextra{The }Common \TeX\ Archive Network}
\acrodef{DDPG}{deep deterministic policy gradient}
\acrodef{DeepLoco}{deep locomotion}
\acrodef{DOI}{Document Object Identifier\acroextra{ (see
    \url{http://doi.org})}}
\acrodef{DPG}{deterministic policy gradient}
\acrodef{DQN}{deep Q-network}
\acrodef{DRL}{deep reinforcement learning}
\acrodef{DYNA}{DYNA}
\acrodef{EOM}{Equations of motion}
\acrodef{EPG}{expected policy gradient}
\acrodef{FDR}{future discounted reward}
\acrodef{FSM}{finite state machine}
\acrodef{GAE}{generalized advantage estimation}
\acrodef{GAIfO}{generative adversarial imitation from observation}
\acrodef{GAN}{generative adversarial network}
\acrodef{GPS}[GPS]{Graduate and Postdoctoral Studies}
\acrodef{HLC}{high-level controller}
\acrodef{HLP}{high-level policy}
\acrodef{HRL}{hierarchical reinforcement learning}
\acrodef{KLD}{Kullback-Leibler divergence}
\acrodef{LLC}{low-level controller}
\acrodef{LLP}{low-level policy}
\acrodef{MARL}{Multi-Agent Reinforcement Learning}
\acrodef{MBAE}{model-based action exploration}
\acrodef{MPC}{model predictive control}
\acrodef{MDP}{Markov Decision Processes}
\acrodef{MSE}{mean squared error}
\acrodef{MultiTasker}{controller that learns multiple tasks at the same time}
\acrodef{Parallel}{randomly initialize controllers and train in parallel}
\acrodef{PD}{proportional derivative}
\acrodef{PDF}{Portable Document Format}
\acrodef{PLAiD}{Progressive Learning and Integration via Distillation}
\acrodef{PPO}{proximal policy optimization}
\acrodef{PTD}{positive temporal difference}
\acrodef{RBF}{radial basis function}
\acrodef{ReLU}{rectified linear unit}
\acrodef{RCS}[RCS]{Revision control system\acroextra{, a software
    tool for tracking changes to a set of files}}
\acrodef{RL}{reinforcement learning}
\acrodef{SGD}{stochastic gradient descent}
\acrodef{Scratch}{randomly initialized controller}
\acrodef{SIMBICON}{SIMple BIped CONtroller}
\acrodef{SMBAE}{stochastic model-based action exploration}
\acrodef{SVG}{stochastic value gradients}
\acrodef{SVM}{support vector machine}
\acrodef{TCN}{time contrastive learning}
\acrodef{TD}{temporal difference}
\acrodef{TL}{transfer learning}
\acrodef{terrainRL}{terrain adaptive locomotion}
\acrodef{TLX}[TLX]{Task Load Index\acroextra{, an instrument for gauging
  the subjective mental workload experienced by a human in performing
  a task}}
\acrodef{TRPO}{trust region policy optimization}
\acrodef{UBC}{University of British Columbia}
\acrodef{UCB}{upper confidence bound}
\acrodef{UI}{user interface}
\acrodef{UML}{Unified Modelling Language\acroextra{, a visual language
    for modelling the structure of software artefacts}}
\acrodef{URDF}{unified robot description format}
\acrodef{URL}{Unique Resource Locator\acroextra{, used to describe a
    means for obtaining some resource on the world wide web}}
\acrodef{W3C}[W3C]{\acroextra{the }World Wide Web Consortium\acroextra{,
    the standards body for web technologies}}    
\acrodef{XML}{Extensible Markup Language}

\title{Is Exploration or Optimization the Problem for Deep Reinforcement Learning?}

\author{ 
    Glen Berseth \\
    Université de Montréal, Mila - Quebec AI Institute, and CIFAR \\
    glen.berseth@mila.quebec
  }
\begin{document}
\maketitle

\setlength\abovecaptionskip{0.1cm}

\begin{abstract}

In the era of deep reinforcement learning, making progress is more complex, as the collected experience must be compressed into a deep model for future exploitation and sampling. Many papers have shown that training a deep learning policy under the changing state and action distribution leads to sub-optimal performance, or even collapse. This naturally leads to the concern that even if the community creates improved exploration algorithms or reward objectives, will those improvements fall on the \textit{deaf ears} of optimization difficulties. This work proposes a new \textit{practical} sub-optimality estimator to determine optimization limitations of deep reinforcement learning algorithms. Through experiments across environments and RL algorithms, it is shown that the difference between the best experience generated is 2-3$\times$ better than the policies' learned performance. This large difference indicates that deep RL methods only exploit half of the good experience they generate. 
\end{abstract}

\section{Introduction}
\label{sec:intro}

What is preventing deep reinforcement learning from solving harder tasks? 
Many papers have shown that training a deep learning policy under the changing state distribution (non-IID) leads to sub-optimal performance~\citep{NikishinSDBC22PrimacyBias,LyleZNPPD23Understandingplas,Dohare2024-tq}. However, at a macro scale, it is not completely clear what causes these issues. Do the network and regularization changes from recent work improve exploration or exploitation, and which of these two issues is the larger concern to be addressed to advance deep RL algorithms? For example, better exploration algorithms can be created, but will the higher value experience fall on the \textit{deaf ears} of the deep network optimization difficulties?

How can we understand if the limited deepRL performance is due to a lack of good exploration or deep network optimization (exploitation)? Normally in RL, to understand if there is a limitation, an oracle is needed to understand \textit{sub-optimality}, how far the algorithm is from being optimal. However, that analysis is with respect to the best policy and aliases both causes of the limitations of either exploration or optimization. Instead, 
consider the example where a person is learning how to build good houses. There are two issues that may prevent the person from \textit{consistently} building a high quality house: (1) they can't \textit{explore} well enough to discover a good design or (2) they can explore well enough to find good designs, but they can't properly \textit{exploit} their experience to replicate those good experience. For deep RL algorithms, which of these two issues is more prevalent? 

To understand if exploration or exploitation is the larger culprit, a method is needed to estimate the \textit{practical} \textit{sub-optimality} between these cases.
This estimator should (1) measure the agent's ability to explore, (2) while also estimating the average performance for the learning policy $\pi^{\theta}$.
While estimating the average policy performance is common, estimating the exploration ability for a policy is not. Extending the house-building metaphor, the idea is to estimate how close the agent ever got to constructing a good home.
Therefore, to realize this estimator, we propose computing the \textit{exploration} value for a policy that is calculated over prior experience, called the \textit{experience optimal policy}.
Using this concept, a new version of sub-optimality can be developed that can compute the difference between the \textit{experience optimal policy} and the learned policy, shown in~\Cref{fig:exp-explotation-gap}. 
If there is a large difference between these two, then the performance is limited by exploitation and optimization (model); however, if the difference between the \textit{experience optimal policy} and the \textit{learned policy} is small, then performance is limited by exploration (data).

The described estimator is used to better understand the reason deepRL algorithms do not solve certain \textit{difficult} tasks. It is found that the limitation of deepRL agents in making progress on difficult tasks is not exploration but often exploitation. 
Therefore, this paper argues that to advance deep reinforcement learning research, further work is needed on optimization for exploitation under non-iid data.
The proposed metric can serve multiple additional purposes. (1) For any RL practitioner, this metric can be used to quickly identify if the limitation in performance is an exploitation or exploration problem so that they can focus their efforts. (2) For the research community, this metric can be used across environments and algorithms to understand the performance of deepRL algorithms better and shed light on the \textit{exploration} vs \textit{exploitation} trade-off on a macro sense, to determine if to increase RL progress the community should be working more on exploitation problems\footnote{This is not a judgement on the exploration community, in fact it is with exploration community in mind this work started so that their amazing research gets the best analysis it can, and great exploration algorithms are not misunderstood due to exploitation problems.}. (3) Showing that including exploration bonuses or scaling network size with RL algorithms increases the \methodName, indicating that optimization becomes a larger issue in that setting.
\Cref{sec:results} provides evidence to showcase these uses of this new view on sub-optimality, and finds for many environments the difference between the \textit{experience optimal policy} and the learned policy to be larger than the difference between the learned policy and the initial policy. These findings suggest a significant exploitation issue and a need for improved optimization methods in RL.

\section{Related Work}
\label{sec:related-work}

Since the first successes of RL and function approximation~\citep {tesauro1995temporal,Mnih2015HumanlevelCT,OpenAI2019-fa}, many recent works have shown great progress on integrating the complexities of deep learning and reinforcement learning~\citep{Hansen2022-za,Hasselt18DRLDeadly}.
Many have studied that certain model classes and loss assumptions make it easier to train more performant deepRL policies~\citep{SchwarzerOCBAC23BBF,Farebrother2024-xr},
deepRL is even used to fine-tune the largest networks to create strong LLMs~\citep{chatgpt}. While deepRL is now being used across a growing number of applications, the broad limitations of current algorithms become less clear.

\paragraph{Deep Reinforcement Learning Training}
The field of methods to explain and improve on the limitations of combining function approximation and reinforcement learning (deepRL) is expanding. Much of the early work consisted of improving value-based methods to overcome training and non-IID data issues in DQN~\citep{Mnih2015HumanlevelCT} and DDPG~\citep{Lillicrap2015DDPG} and stochasticity~\citep{TRPO,SchulmanWDRK17PPO}. Recent adaptations improve over the initial algorithms that struggle with overestimation~\citep{HasseltGS16DoubleDQN,BellemareDMC51,HesselMHSODHPAS18RAINBOW} or improving critic estimation~\citep{Fujimoto2018TD3,HaarnojaZAL18SAC,LanPFW20maxminq,KuznetsovSGV20tqc,ChenWZR21redq}.
The challenges in the space of learning policy are based on an unstable mix of function approximation, bootstrapping, and off-policy learning, called the Deadly Triad in DRL~\citep{Hasselt18DRLDeadly,Achiam2019Towardscharacterizing}. Many works focus on parts of triad, including: stabilizing effect of target network~\citep{ZhangYW21Break,Chen2022Targetnetwork,Alexandre2022bridging}, difficulty of experience replay~\citep{SchaulQAS15PER,Kumar0L20Discor,OstrovskiCD21Passive}, over-generalization~\citep{GhiassianRLW20ImprovingbyBreaking,PanBW21FuzzyTiling,YangAA22OvercomingSpectral}, representations in DRL~\citep{Zhang21Learning,LiTZHLWMW22HyAR,tang2022PeVFA}, off-policy correction~\citep{Nachum19AlgaeDice,ZhangD0S20GenDICE,Lee21OptiDICE}, interference~\citep{CobbeHKS21PPG,RaileanuF21Decouplingvp,BengioPP20Interference} and architecture choices~\citep{OtaOJMN20Canincreasing}.

\paragraph{DeepRL Exploration Methods} 
On top of the above training stability improvements is the desire to improve exploration by providing the agent with better signal to encourage exploration beyond just the extrinsic reward.
These intrinsic rewards often compute some measure of state visitation or mutual information using a separate online learnt model. Count-based methods (curiosity) are early examples that encourage agents to cover a larger state space~\citep{bellemare2016unifying, ostrovski2017countbased, tang2017exploration,burda2018exploration}, but they do not scale well to large state spaces. 
Several works \citep{pathak2017curiositydriven, badia2020up, zhang2020bebold, zhang2021noveld} have built on curiosity frameworks to improve training and learning. However, it is not known how well RL algorithms will be able to learn from the additional experience.

\paragraph{DeepRL Scaling Methods} Given the significant gains of using large models on many supervised learning problems, the RL community has been studying how to achieve similar gains from scale, but deep RL performance often drops when larger networks are used ~\citep{SchwarzerOCBAC23BBF,tang2024improving}. 
Recent works focus on network structure changes to avoid divergence and collapse, using normalization layers~\citep{BRO,Lyle2024-bq}, regularization~\citep{NikishinSDBC22PrimacyBias,SchwarzerOCBAC23BBF,Galashov2024-gy} or optimization adjustments~\citep{Lyle2024-bq}.
The goal in these prior works is to understand and improve performance when larger networks are used, but these papers are often limited to recovering prior performance, not understanding where RL in general is missing potential.

\section{Background}
\label{sec:background}

In this section, a very brief review of the fundamental background of the proposed method is provided. \ac{RL} is formulated within the framework of an \ac{MDP} where at every time step $t$, the world (including the agent) exists in a state $ \bs_{t} 
\in \states $, where the agent is able to perform actions $ \ba_{t} \in 
\actions $. The action to take is determined according to a policy $ \pi(\ba_{t}|\bs_{t})$ which
results in a new state $ \bs_{t+1} \in \states $  and reward $\reward_{t} = R(\bs_t, \ba_t)$ according to the transition 
probability function $ P(\bs_{t+1} | \bs_{t}, \ba_{t}) $. 
The policy is optimized to maximize the future discounted reward
$
  \expectation_{\reward_{0}, ..., \reward_T} \left[ \sum_{t=0}^{T} \gamma^t \reward_{t} \right]$,
\noindent where $ T $ is the max time horizon, and $ \discountFactor $ is the 
discount factor.
The formulation above generalizes to continuous states and actions. There are multiple RL algorithms that can be used to optimize the above objective. This work uses two of the most popular algorithms DQN~\citep{Mnih2015HumanlevelCT} and PPO~\citep{SchulmanWDRK17PPO} to frame the challenges with optimizing and exploration.

\paragraph{Policy Gradient Definitions}

To discuss the difference between policy performance and estimators, it is useful to define the state visitation distribution $d_{s_0}^\pi(s)$ for a policy:
\begin{equation}
d_{s_0}^\pi(s) := (1 - \gamma) \sum_{t=0}^\infty \gamma^t \Pr(s_t = s | s_0),
\end{equation}
where $\Pr^\pi(s_t = s | s_0)$ is the probability of the policy $\pi$ visiting the future state $s_t$ when starting from $s_0$.
The policy gradient can be written in the form
\begin{equation}
\nabla_\theta V^{\pi_\theta}(s_0) = \frac{1}{1 - \gamma} \mathbb{E}_{s \sim d_{s_0}^{\pi_\theta}} \mathbb{E}_{a \sim \pi_\theta(\cdot|s)} \left[ \nabla_\theta \log \pi_\theta(a|s) Q^{\pi_\theta}(s, a) \right].
\end{equation}
Then we can write out the \textbf{performance difference lemma}~\citep{kakade2002approximately}  between two policies as
\begin{equation}
\label{eq:performance-diff}
V^{\pi'}(s_0) - V^{\pi}(s_0) = \frac{1}{1 - \gamma} \mathbb{E}_{s \sim d_{s_0}^\pi} \mathbb{E}_{a \sim \pi(\cdot|s)} \left[ A^{\pi'}(s, a) \right].
\end{equation}
Where $A^{\pi'}(s, a)$ is the advantage of policy $\pi'$.

\section{Is Exploration or Exploitation the Issue for DeepRL?}
\label{sec:method}

Often, learning agents are concerned with the exploration vs exploitation trade-off. This trade-off is a helpful lens for discussing an agent's choices at a particular state $\bs_t$, but this single state view focuses on \textit{exploitation} as either: a type of greedy action selection, sampling from a learned policy, or utilizing a world model. However, in the age of deep learning and ever increasing model and data sizes, that lens misses the bigger picture of the definition \textit{exploitation is making use of prior experience}, in that for each of these types of exploitation, there is a deep network $\theta$ optimization process over some experience $\data$ that is imperfect. However, it is not clear if the difference is from data distribution issues~\citep{OstrovskiCD21Passive} or optimization~\citep{Lyle2024-bq}.  
To improve the understanding of the limitations of RL with function approximation (deepRL), we introduce estimators to quantify the difference between a policy's data-generating process (exploration/data) and its ability to learn from that data (exploitation/model).

\begin{figure*}[htb]
\centering
\subcaptionbox{\label{fig:exp-explotation-gap} Example exploitation sub-optimality difference}{ \includegraphics[trim={0.0cm 0.0cm 0.0cm 0.0cm},clip,height=0.17\textheight]{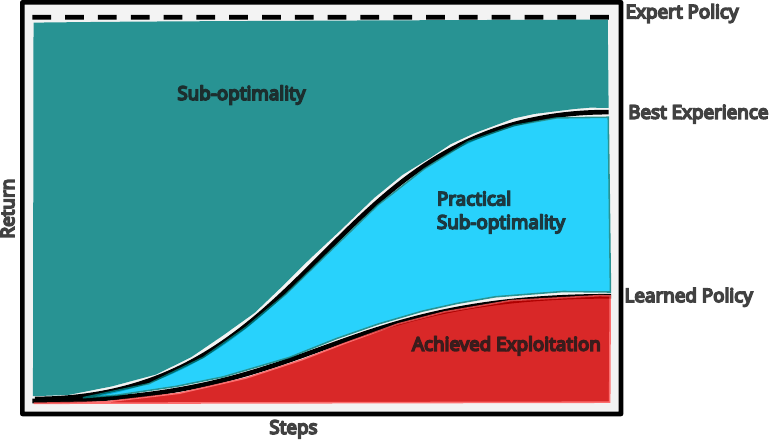} }
\subcaptionbox{\label{fig:minAtar-spaceInvaders-single-task} MinAtar space invaders DQN}{ \includegraphics[trim={0.0cm 0.0cm 0.0cm 0.0cm},clip,height=0.17\textheight]{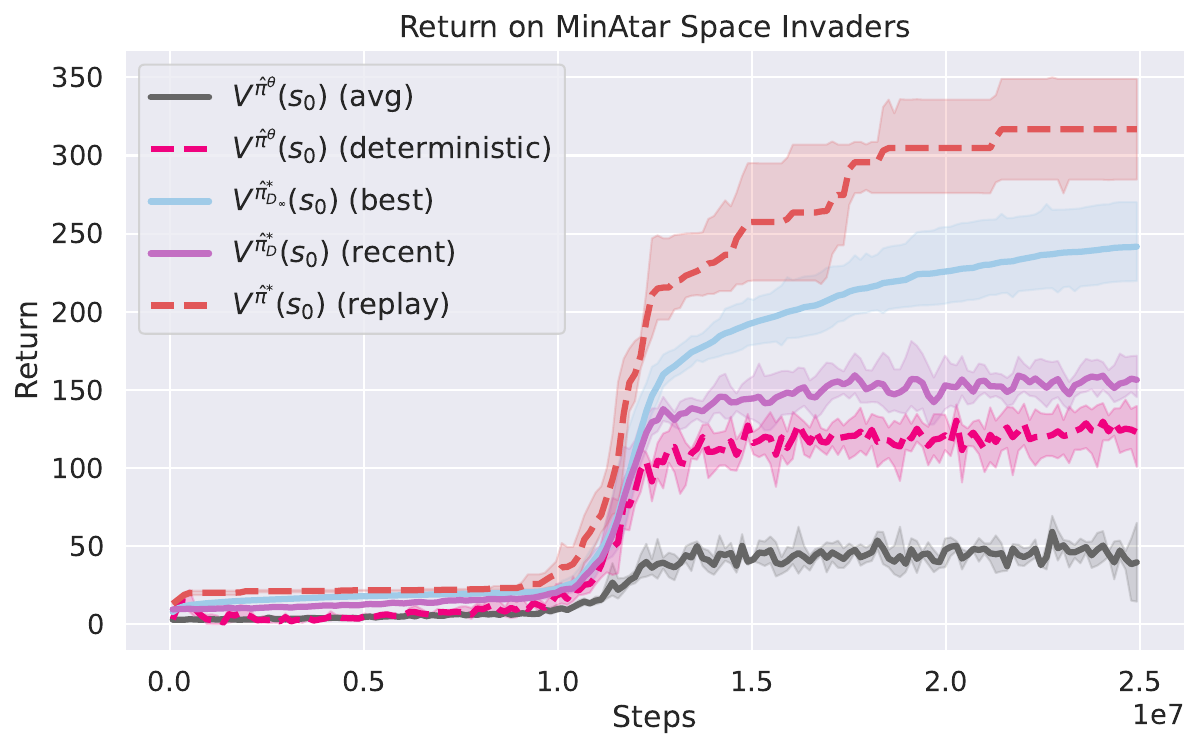}}
\caption{
Left: Diagram of the \methodName = Best Experience - Learned Policy. On the right are results computing this exploitation gap as the difference between \topBest and \policyAverage in MinAtar SpaceInvaders.
}
\label{fig:global-optimality-diagram}
\vspace{-0.25cm}
\end{figure*}
In \Cref{fig:global-optimality-diagram} we show the conceptual version of studying this exploration vs exploitation problem, where the typical learning graph is now split into three sections: the performance of the average policy $\pi^{\theta}$ from \textit{achieved exploitation} (red), which measures what that policy has learned, 
the potential performance, indicated by the optimal policy $\pi^*$, and a new estimator we call the \textit{experience optimal} policy $\hat{\pi}^*$. The challenge is that $\pi^{\theta}$ can be arbitrarily bad compared to $\pi^*$, and normally it is not clear if the performance difference (\Cref{eq:performance-diff}) is because the agent is not exploring well (\topBest $<<$ \expert and \policyAverage $<<$ \expert) or just not exploiting well (\policyAverage $<<$ \expert). 
This analysis can be particularly useful for evaluating exploration-focused algorithms. When evaluating the performance of a method, if only \policyAverage is considered, the analysis can miss the fact that the method is generating higher value experiences \topBest, but the policy is not able to exploit them into $\theta$ properly. Therefore, to better understand reinforcement learning limitations, we introduce a new estimator for $\hat{\pi}^*$ to measure practical sub-optimality.

\paragraph{How to measure practical sub-optimality}
The optimal policy is defined as the policy that selects the best action at every state~\citep{bellman1954theory}. Sub-optimality measures the difference between a policy's value $V^{\pi}(s)$ with respect to an optimal policy $\pi^*$ with the value function $V^{\pi^*}(s)$. However, if the policy $\pi$ can not explore optimally, using $\pi^*$ is not very informative.
Therefore, in addition to the theoretical optimal policy, we introduce the \textit{experience optimal} policy $\hat{\pi}^*$ to represent the best policy the agent can achieve given the experience collected during training. 
If the environment is deterministic and the agent keeps a buffer of all prior experience $D^{\infty}$, then,
\begin{equation}
\label{eq:pi-max}
 \hat{\pi}^{*} = \argmax_{<a_0, \ldots, a_t> \in D^{\infty}} \sum_{t=0}^{T} r(a_t, s_t)
\end{equation}
This policy can also be understood as deterministically replaying the highest value sequence of actions $<a_0, \ldots, a_t>$ in the experience memory. This policy can be used to compute a new difference as the \textit{exploitation sub-optimality} of the form \topBest $-$ \policyAverage.

Most empirical works use $V^{\pi^\theta}(s_0)$ for comparing across algorithms to understand which algorithm performs the best on a set of tasks. While this information is helpful and enables the community to make steps forward in terms of performance, it does not provide information on why one algorithm is better than another. Consider the example where there are two algorithms A and B, algorithm A generates higher-value experience, but is not able to exploit them, and B does not generate higher-value experience, similar to its policy, but has been able to exploit that data well. Both A and B can have the same value $V^{\pi^A}(s_0) = V^{\pi^B}(s_0)$. Is A or B a better RL algorithm? In this work, we propose that B is the better algorithm, as it can properly exploit generated data. If the experience were equal, algorithm B would see the same experience as algorithm A, then B would result in better performance and have a smaller \methodName.

\subsection{Softer Expert Estimators}
While \Cref{eq:pi-max} is a clear definition for computing an estimate of an optimal policy, it works best when the environment is completely deterministic. For example, the sequence of actions $a_0, \ldots, a_t$ can be replayed in the environment to reproduce \topBest, however, this restrictive definition is less useful for non-deterministic environments where it is impossible for a stochastic policy/trajectory to outperform a deterministic one. Therefore, two additional methods are introduced to estimate the \textit{potential} for the policy to learn from the data.

For the analysis, two versions of \topBest are introduced to approximate the performance on the \textit{best experience}. For stochastic environments, the first version the best stocastic policy from the collected experience as top $5\%$ of experience generated by the agent \topBestFive, where $D_{\infty}$ is all the experience collected by the agent. The second is the \textit{recent} top $5\%$ of data \topBestLocal in the replay buffer $D$. To estimate the value function $V(s_0)$ from data, the sum of rewards the agent achieves in the environment, or the return, is used. The true value is computed using this function:
\begin{equation}
\label{eq:pi-max-top-k}
V^{ \hat{\pi}^* }(s_0) = \frac{1}{k}\sum\limits_{\tau \in D_{0:k}} \sum\limits_{a_t,s_t \in \tau} R(a_t,s_t)
\end{equation}
Where $k$ is equal to $\frac{1}{20} \times |D|$ and $D$ is sorted with the highest value trajectory starting at index $0$.

The best \textit{ever} and \textit{recent} estimators both have their own reasoning. The best \textit{ever} experience \topBestFive is a measure of how good the agent is at exploiting the best experience it ever generated. This notion is rather strong and difficult for any RL algorithm to match, as the agent may not currently have access to that experience for optimization, but it is a notion of lifetime achievement and represents a possible high-value policy and trajectory the agent could generate again. The \textit{recent} best experience \topBestLocal is a measure of the agent's ability to learn to match the best of the recent experience it has access to and can use for optimization. The \textit{recent} notion can be more fair as it is possible for an agent to train on that experience to improve its performance actively, but as will be shown in \Cref{seq:results-per-task}, RL algorithms also struggle to match this performance.

\subsection{For RL Algorithms}
\label{sec:aggregate-alg}
The above estimators can be used to understand the \methodName of an algorithm on an environment. That information is useful, but it does not speak about an algorithm holistically. For example, we may have the question, \textit{how much does an algorithm suffer from exploitation limitations} and \textit{which algorithms are the best at exploiting their generated data}?
This information is paramount for the community to understand better where there is a larger benefit from time spent on research and development. To compute this information \textit{across environments}, the estimator will need to be aggregated and normalized across environments.

To compute this aggregate estimator the upper bound from \topBest can be used in place of the less accurate and often overestimated the \textit{optimalality gap} from \textit{rliable}~\citep{Agarwal2021-xx}. The gap computed using the proposed metric is also relative to the data the agent has generated, which can provide more rich signal than comparing the performance to some potential unattainable perfect agent. For example, when the optimal performance is not know a heuristic is often used to compute the return for the optimal policy by taking the max possible reward $r_{\text{max}}$ and multiplying this by the inverse of the discount factor $V^{\pi^*}(s_0) \approx r_{\text{max}} * \frac{1}{1-\gamma}$. Instead, the proposed metric can be used to calculate the \methodName for an RL algorithm as:
\begin{equation}
    \label{eq:gap-aggregate}
    \frac{1}{|\mathcal{T}|} \sum\limits_{m \in \mathcal{T}} (V^{\hat{\pi}^*}_m(s_0) - V^{\pi^\theta}_m(s_0)) / (V^{\hat{\pi}^*}_m(s_0) - V^{\pi^0}_m(s_0) ).
\end{equation}
Where $m$ is some task or environment. This metric is used in~\Cref{seq:results-per-alg} to compare the aggregate weaknesses across RL algorithms.

\paragraph{Implementation Details} It is difficult to compute a general \methodName for any type of RL algorithm. On-policy algorithms do not keep around histories of recent data for evaluation, and off-policy algorithms don't track returns as they often use Q-functions for learning directly from rewards. To facilitate the tracking of these statistics, we introduce a wrapper that can be introduced into the RL algorithm code to track every reward, return, and end of episode. This wrapper is also used to compute the \textit{best} \topBestFive and the \textit{recent} \topBestLocal version.

\section{Experimental Results}
\label{sec:results}

In this section, the ability of \methodName for diagnosing learning issues is evaluated. This usefulness is determined in multiple ways: (1, \Cref{seq:results-per-task}) As a metric to determine the limitations of current RL algorithms on specific environments, (2) how recent methods for exploration or scaling increase or reduce the \methodName, and (3) The overall limitations of RL algorithms and if more exploration or exploitation is needed to improve performance over difficult/unsolved tasks.

Two popular RL algorithms are used for evaluation. First PPO~\citep{SchulmanWDRK17PPO} is a common on-policy algorithm used for various problems, known for its ease of implementation and use. The other algorithm is DQN~\citep{Mnih2015HumanlevelCT}, which is a popular RL algorithm for environments with discrete actions. These two algorithms cover the most common use cases for RL.

A selection of evaluation environments is included to cover a diverse range of the RL landscape. This diverse selection is important to understand better the \methodName there needs to be a difference between the generated data and the final policy's performance. Therefore, we focus on including experimental results on environments that are \textit{difficult}. These difficult environments include using MinAtar~\citep{Young2019-dn} and Atari~\citep{ale,atariFive} \textbf{SpaceInvaders}, \textbf{Asterix}, \textbf{LunarLander}, \textbf{Montezumas Revenge}, \textbf{Craftax}, and the \textbf{AtariFive}~\citep{atariFive}. We also include \textbf{Walker2d},  \textbf{HalfCheeta}, \textbf{Humanoid} as continuous action environments that are easier, and as will be shown, have little \methodName.

\begin{figure*}[htb]
\centering
\subcaptionbox{\label{fig:halfCheetah-env} HalfCheetah}{ \includegraphics[trim={0.0cm 0.0cm 0.0cm 0.0cm},clip,height=0.15\textheight]{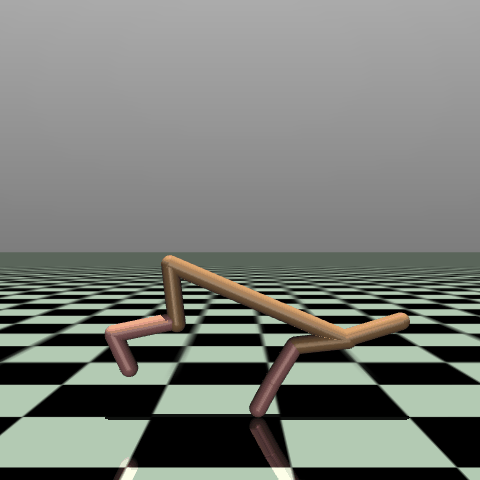}}
\subcaptionbox{\label{fig:breakout-env} MinAtar}{ \includegraphics[trim={0.0cm 0.0cm 0.0cm 0.0cm},clip,height=0.15\textheight]{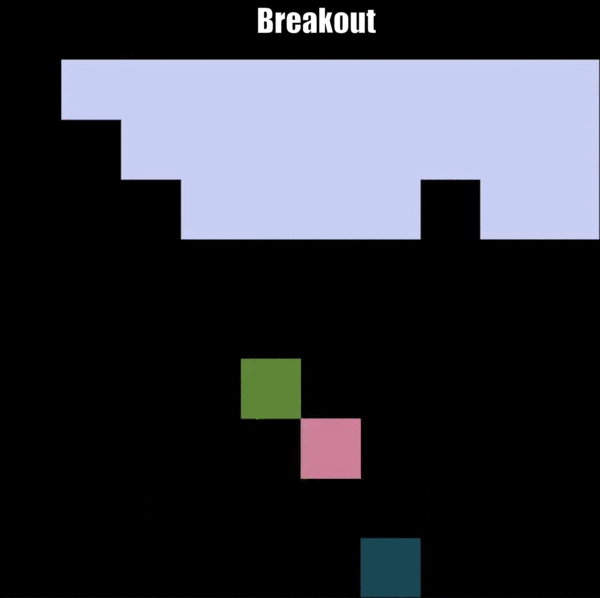}}
\subcaptionbox{\label{fig:MR-env} MontezumaRevenge}{ \includegraphics[trim={0.0cm 0.0cm 0.0cm 0.0cm},clip,height=0.15\textheight]{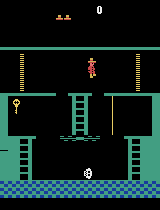}}
\subcaptionbox{\label{fig:space-invaders-env} Atari}{ \includegraphics[trim={0.0cm 0.0cm 0.0cm 0.0cm},clip,height=0.15\textheight]{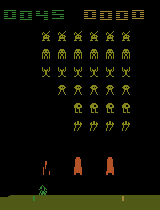}} 
\caption{
Evaluation environments include examples from Mujoco, MinAtar, and Atari.
}
\label{fig:exp-envs}
\vspace{-0.25cm}
\end{figure*}

To measure performance, we will look at the \methodName discussed in the previous section. In addition, the average return during learning is used to verify that the agents are learning, ensuring that the reason for the lack of \methodName is not due to the agent's inability to learn. All experiments are conducted over $4$ random seeds.

\subsection{Per Task Sub-optimality}
\label{seq:results-per-task}

\begin{figure*}[htb]
\centering
\subcaptionbox{\label{fig:halfCheetah} PPO HalfCheetah}{ \includegraphics[trim={0.0cm 0.0cm 0.0cm 0.0cm},clip,width=0.31\linewidth]{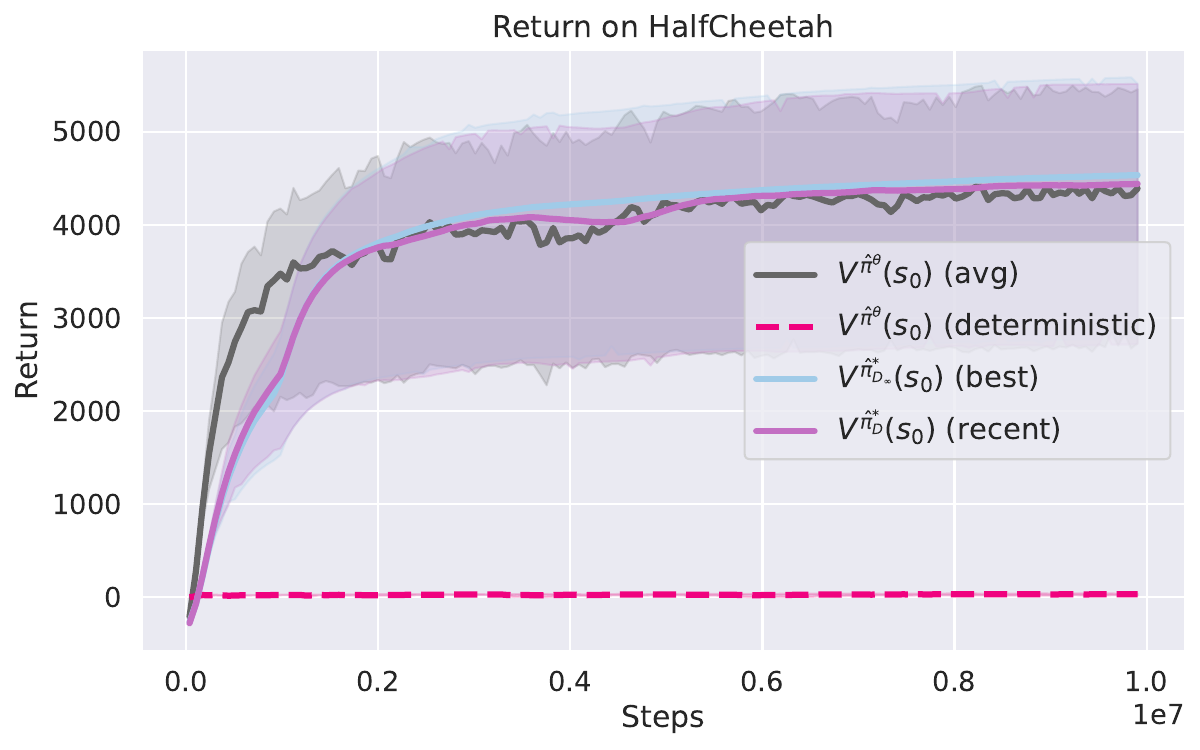}}
\subcaptionbox{\label{fig:mr-ppo-compare} PPO Montezums Revence}{ \includegraphics[trim={0.0cm 0.0cm 0.0cm 0.0cm},clip,width=0.31\linewidth]{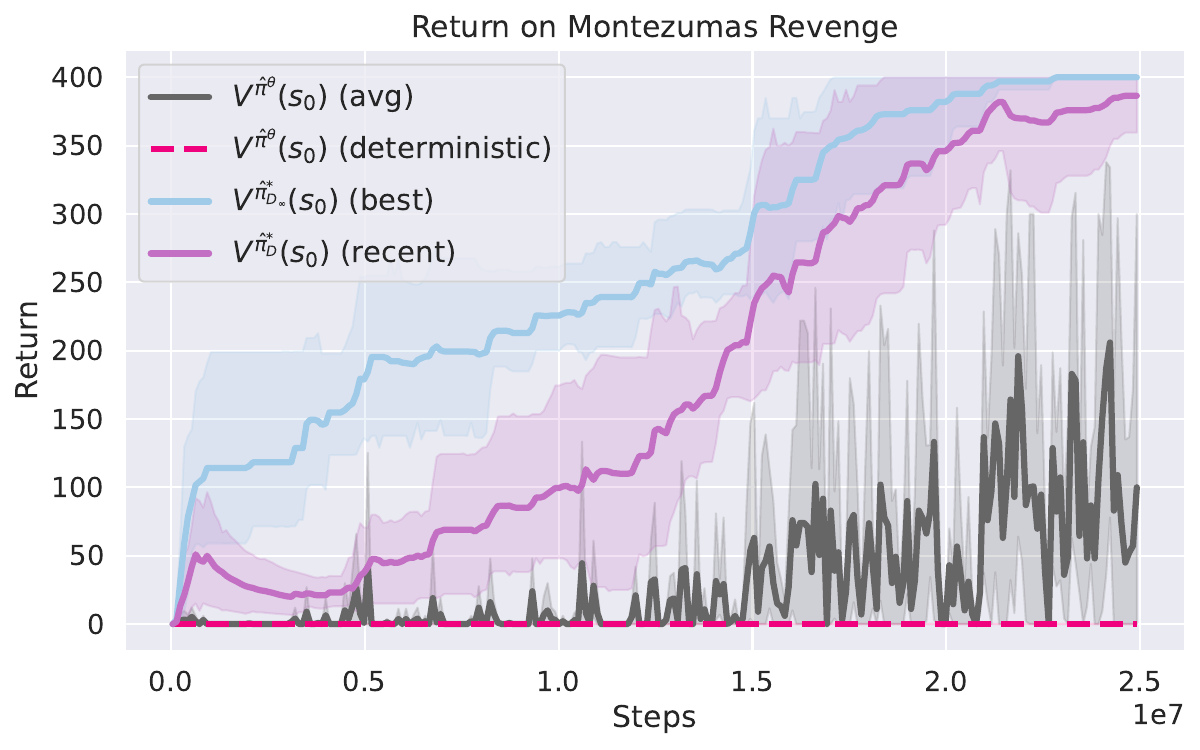}}
\subcaptionbox{\label{fig:minAtar-breakout-single-task} MinAtar/Breakout DQN }{ \includegraphics[trim={0.0cm 0.0cm 0.0cm 0.0cm},clip,width=0.31\linewidth]{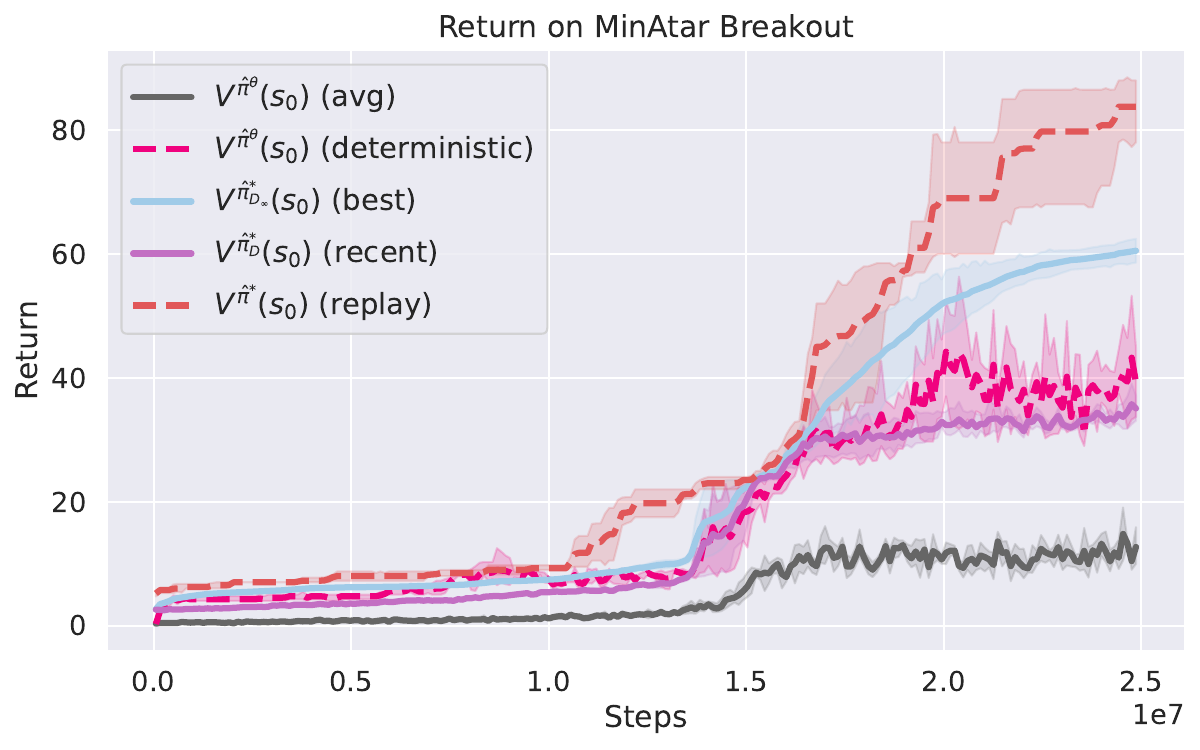}} \\
\subcaptionbox{\label{fig:Atari-NameThisGame-DQN} NameThisGame DQN}{ \includegraphics[trim={0.0cm 0.0cm 0.0cm 0.0cm},clip,width=0.32\linewidth]{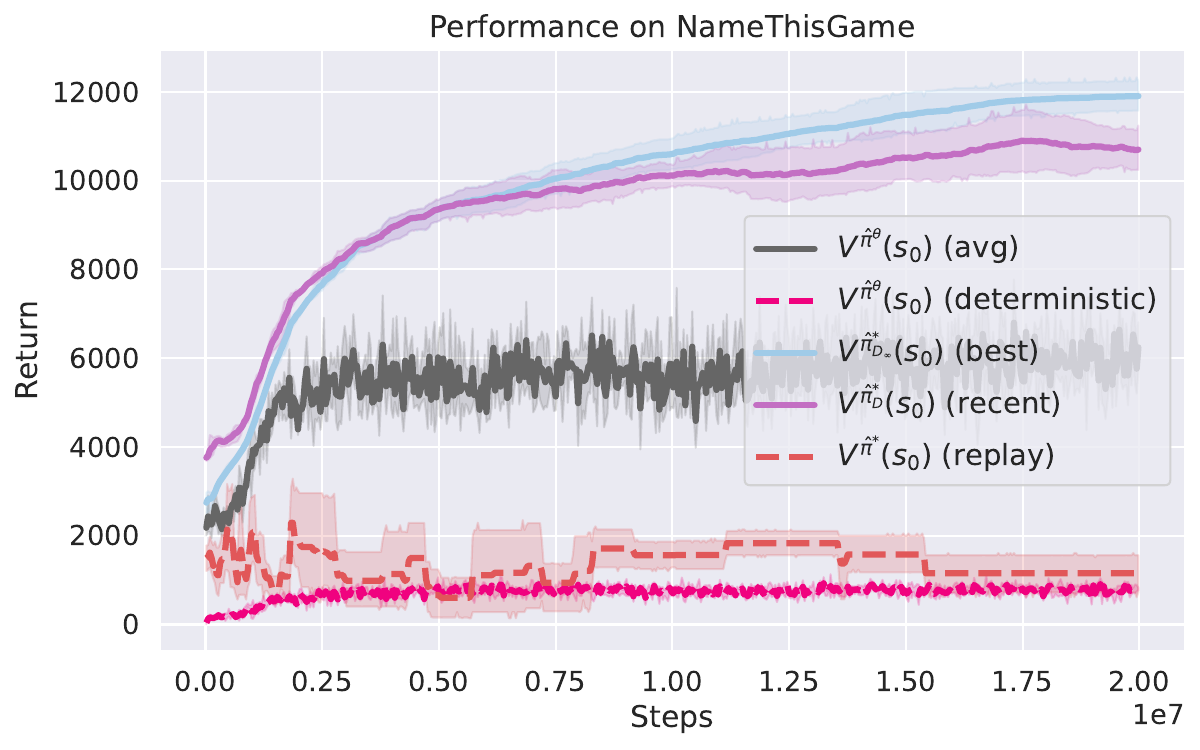}}
\subcaptionbox{\label{fig:Atari-BattleZone-DQN} BattleZone DQN}{ \includegraphics[trim={0.0cm 0.0cm 0.0cm 0.0cm},clip,width=0.32\linewidth]{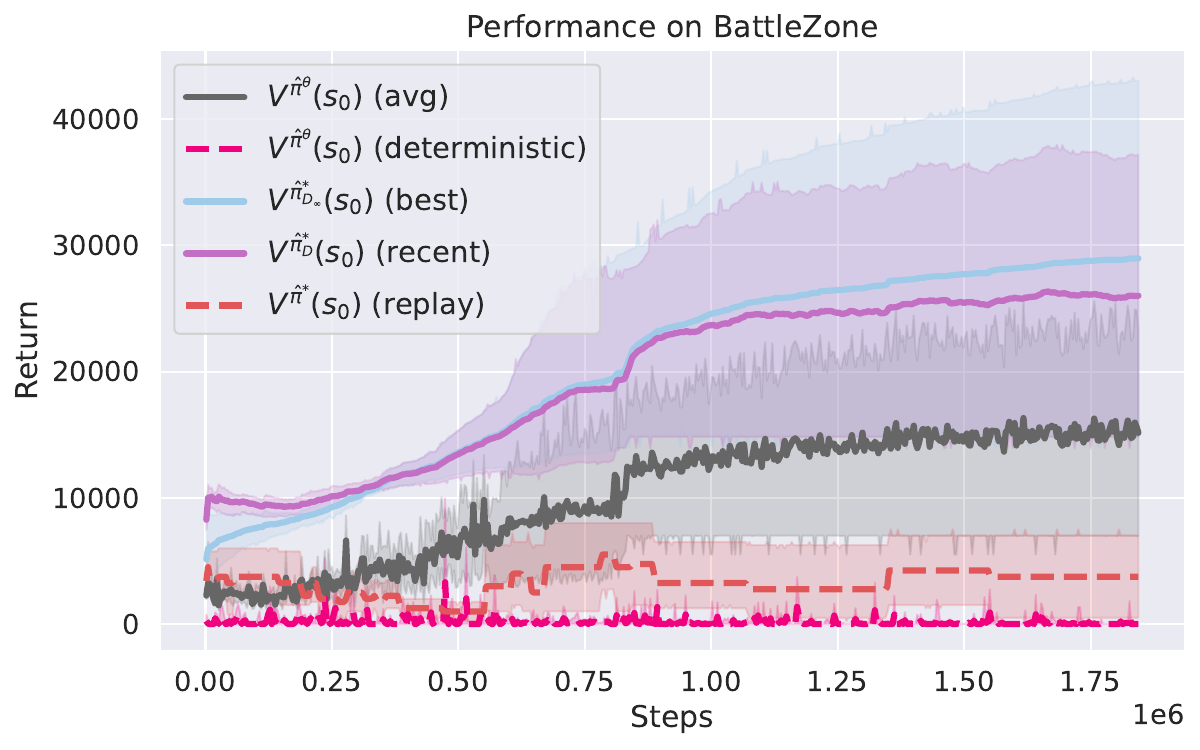}}
\subcaptionbox{\label{fig:Asterix-DQN} Asterix DQN}{ \includegraphics[trim={0.0cm 0.0cm 0.0cm 0.0cm},clip,width=0.32\linewidth]{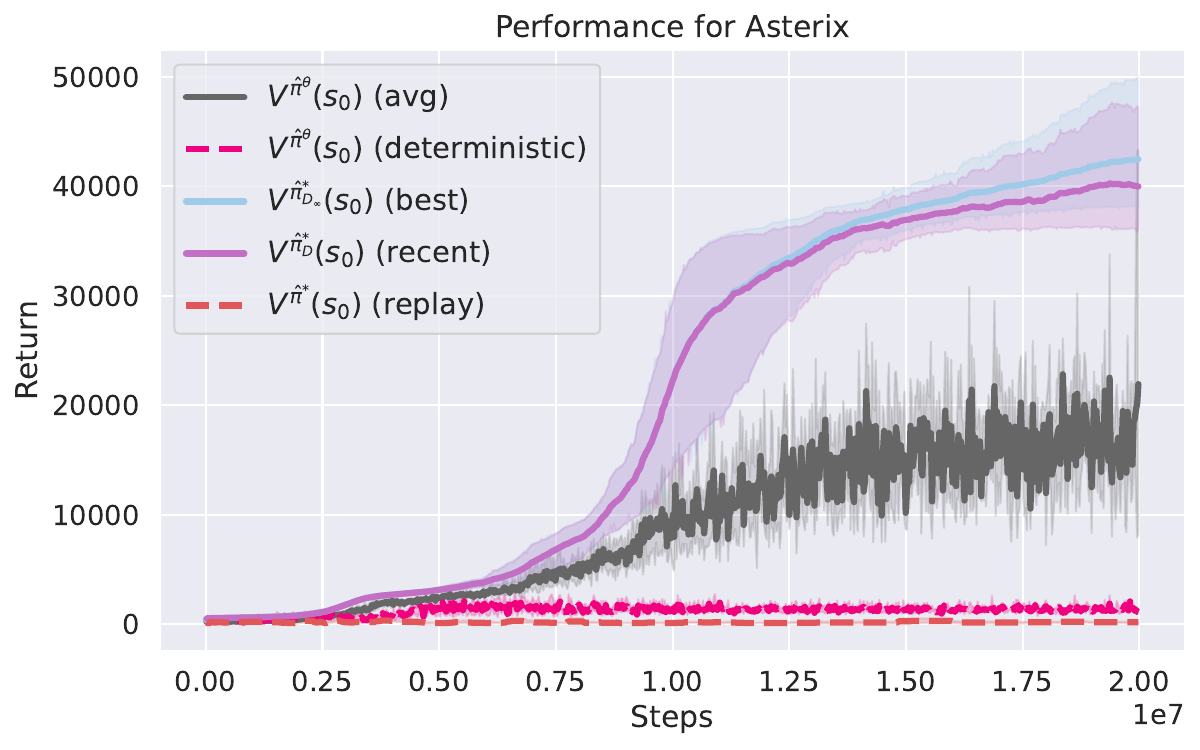}}
\caption{
Comparisons of different measures for global optimality and the learned policy $\pi^\theta$. For environments with more complex exploration, such as Montezuma's Revenge, Breakout and SpaceInvaders, there is a large exploitation gap between \topBestFive and \policyAverage.
}
\label{fig:all-diff}
\vspace{-0.25cm}
\end{figure*}

In this section, we can study which tasks express types of this \methodName, indicating a need for improvements in optimization over exploration. The first question (1) is whether tasks exhibit this type of gap, or if all tasks can be solved, or if policies can properly exploit the experience. In \Cref {fig:halfCheetah}, we can see that for \textbf{HalfCheetah} there is little difference between \topBestFive, \topBestLocal, and \policyAverage, even though a high return is achieved; however, it is well known that \textbf{HalfCheetah} is no longer a difficult task for common RL algorithms. We can also see that the deterministic $\hat{\pi}^*$  poorly estimates the best performance in this non-deterministic environment and instead the softer versions work well.
Examining tasks that are well-known to be difficult exploration problems reveals a different story. After training DQN on \textbf{Montezuma's Revenge} (\Cref{fig:mr-ppo-compare}), there is a surprisingly large gap where \policyAverage is noisy and near zero, yet the policy does generate many high-value trajectories, indicated by both a large difference between \topBestFive and \topBestLocal, but is not able to learn from these. These higher value trajectories are not rare. The \topBestLocal line indicates that the policy, aside from a few spikes, is far from the best $5\%$ of better experiences. We find similar results for many other environments and algorithms shown in \Cref{fig:all-diff}.

The \methodName may be an overestimate of true policy performance. To address this issue, we perform a pure analysis with a set of completely deterministic environments in \Cref{fig:minAtar-spaceInvaders-single-task} and \Cref{fig:minAtar-breakout-single-task}. Because these environments are \textit{deterministic} settings, it is possible to compute a true \topBest which is equal to the best single trajectory ever discovered. 
This best single trajectory is visualized as \topBest, where the policy for \topBest is $a_0, \ldots, a_t$, which is replayed to visualize the score and indicate that to reach this performance, the policy $\theta$ needs to exploit this data well enough to reach that score. 
As can be seen, \topBest $ > $ \topBestFive $ > $ \policyAverage, which indicates that \topBestFive may be slightly lower than the best performance, yet these trained policies struggle to produce the behaviour, indicating that often performance is limited by a lack of exploitation.

Last, to better understand these estimators for stochastic and deterministic settings, it is important to compare deterministic vs stochastic policy performance; in this case, the stochasticity added to the policy is causing a larger difference when the policy has learned a high-value behaviour. For PPO on continuous environments, this is equivalent to taking the mean of the policy, and for a discrete policy, the $\arg\max_a Q(s_t, a)$ is used. In \Cref{fig:halfCheetah} the deterministic policy does poorly, this is likely because the agent quickly reaches states that are out of distribution, causing the agent to fail. Similar is true for \textbf{Montezuma's Revenge} with PPO. However, for MinAtar/Breakout and SpaceInvaders, the $\epsilon$-greedy exploration of DQN knocks the policy off high-value paths, and the deterministic policy does well, even approaching \topBestFive for MinAtar/Breakout. We also see in \Cref{fig:minAtar-spaceInvaders-single-task} and many other results that the difference does not decrease with additional training, indicating the gap is not the result of needing more experience or updates, but more significant changes to improve exploitation and optimization for deep learning.

\subsection{Sub-optimality when adding exploration}
\label{seq:results-exploration}
This section asks the question \textit{does adding exploration objectives increase the difference and therefore aggravate the optimization challenges}. This is analyzed by adding common exploration bonuses to the RL algorithms, RND~\citep{burda2018rnd}. RND applies additional rewards to the extrinsic reward, encouraging the agent to explore a wider distribution of states, which should allow the agent to discover new, higher-reward states. These higher reward states should lead to larger returns, and if the algorithm is not effectively exploiting these rewards, a greater difference will result.

\begin{figure*}[htb]
\centering
\subcaptionbox{\label{fig:global-optimality-explorebonus-MR} PPO MontezumasRevenge}{ \includegraphics[trim={0.0cm 0.0cm 0.0cm 0.0cm},clip,width=0.32\linewidth]{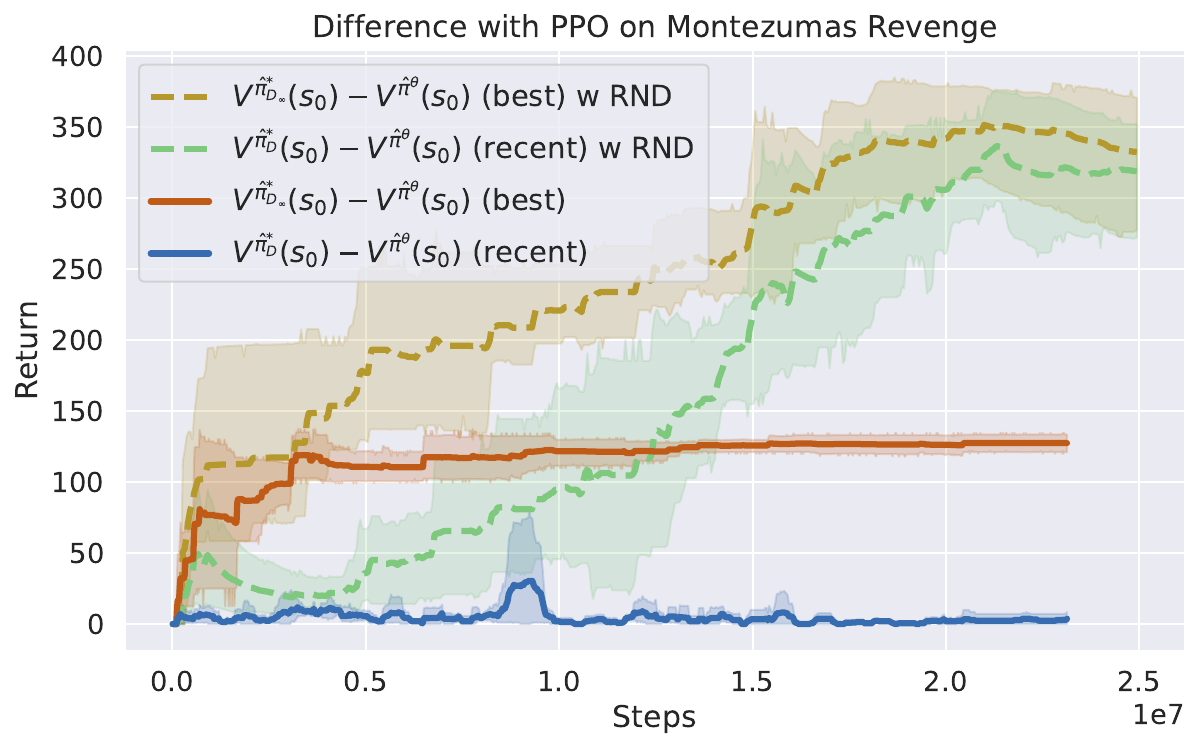}}
\subcaptionbox{\label{fig:global-optimality-explorebonus-SI} PPO SpaceInvaders}{ \includegraphics[trim={0.0cm 0.0cm 0.0cm 0.0cm},clip,width=0.32\linewidth]{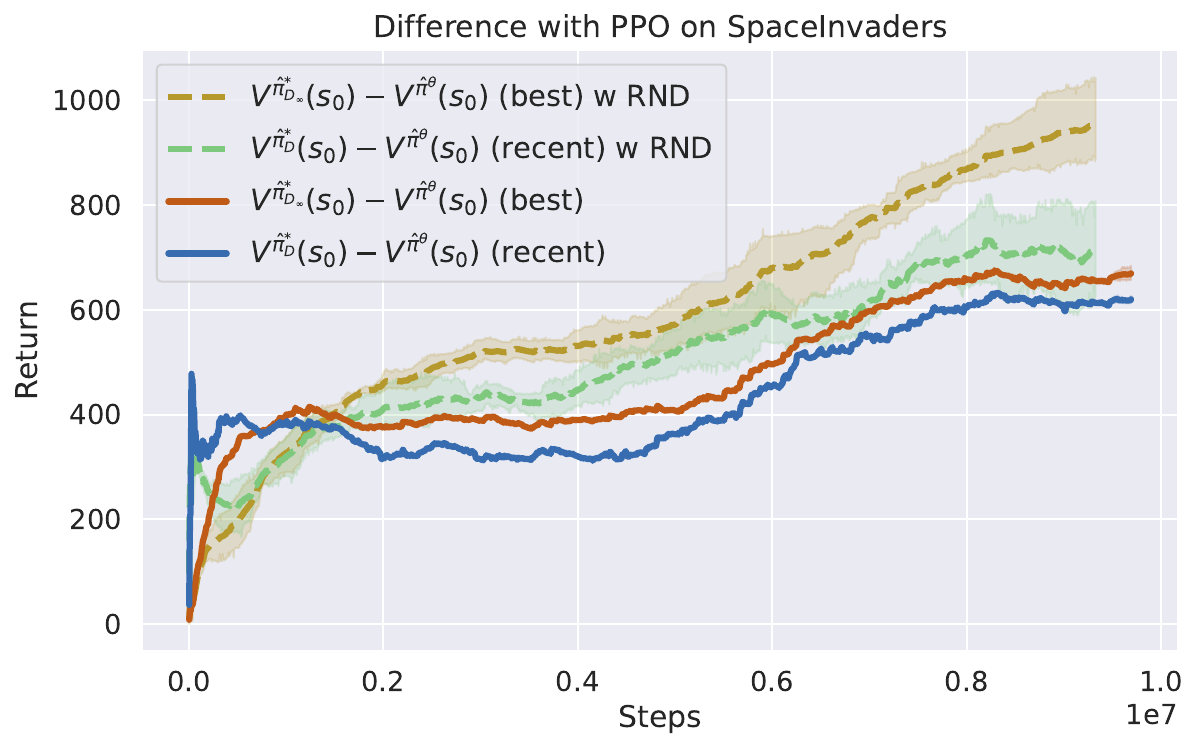}}
\subcaptionbox{\label{fig:global-optimality-explorebonus-Craftax} PPO on Craftax}{ \includegraphics[trim={0.0cm 0.0cm 0.0cm 0.0cm},clip,width=0.32\linewidth]{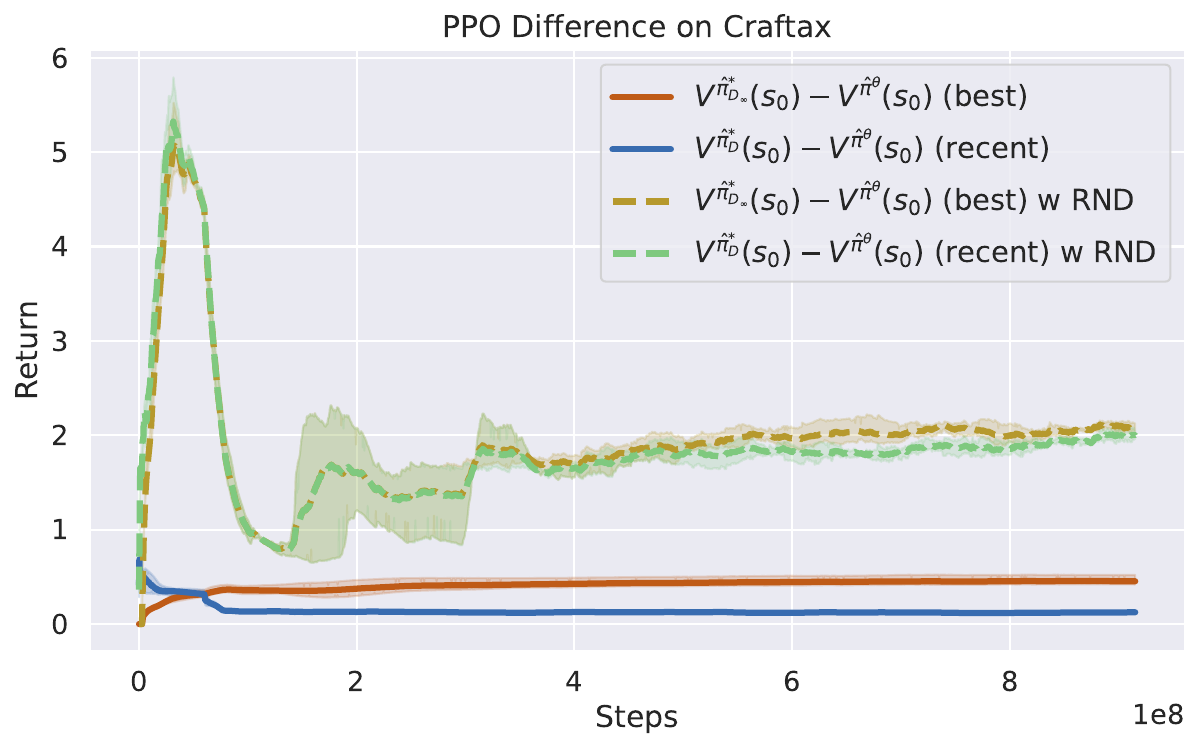}}
\caption{
Comparisons of \methodName for best and recent performance compared to the average using~\Cref{eq:performance-diff} with and without adding \textbf{RND}. These results show that with the addition of \textbf{RND}, the difference increases, indicating that adding exploration objectives is a double-edged sword, better exploration but more difficult exploitation.
}
\label{fig:global-optimality-explorebonus}
\vspace{-0.25cm}
\end{figure*}

\Cref{fig:global-optimality-explorebonus} provides the results of the analysis of \methodName estimating~\Cref{eq:performance-diff} compared with and without using RND. As we can see, the addition of RND improves the returns for DQN and PPO 
. However, the \textit{difference} is also increased, indicating that as exploration is increased, so too are the issues of exploitation of experience in deep RL. This is an undesirable situation; as the agent improves its exploration, it actually learns less from the experience overall due to optimization issues.

\subsection{Sub-optimality when Scaling Networks}
\label{seq:results-scaling}

Many recent reinforcement learning works are discovering improved algorithms' performance based on scaling networks~\citep{SchwarzerOCBAC23BBF,LyleZNPPD23Understandingplas,Obando-Ceron2024-el,BRO,tang2024improving}. 
Are the challenges from scaling just optimization issues, or are these models also struggling to scale because the types of narrow distributions produced by larger models limit exploration? Two experiments were performed to investigate this question with networks of different sizes. First across Atari environments \textbf{BattleZone} and \textbf{NameThisGame} from the Atari-5 group~\citep{atariFive} that is representative of the Full Atari Benchmark, and then across \textbf{HalfCheeta}. For the Atari environments, a comparison is made between training a policy that uses the normal C-51 type network with a 3-layer CNN and using a ResNet18. For the \textbf{HalfCheetah} environment, different numbers of layers are used between $4$ and $256$.

\begin{figure*}[htb]
\centering
\subcaptionbox{\label{fig:global-optimality-explorebonus-BAttlezone} DQN BattleZone}{ \includegraphics[trim={0.0cm 0.0cm 0.0cm 0.0cm},clip,width=0.32\linewidth]{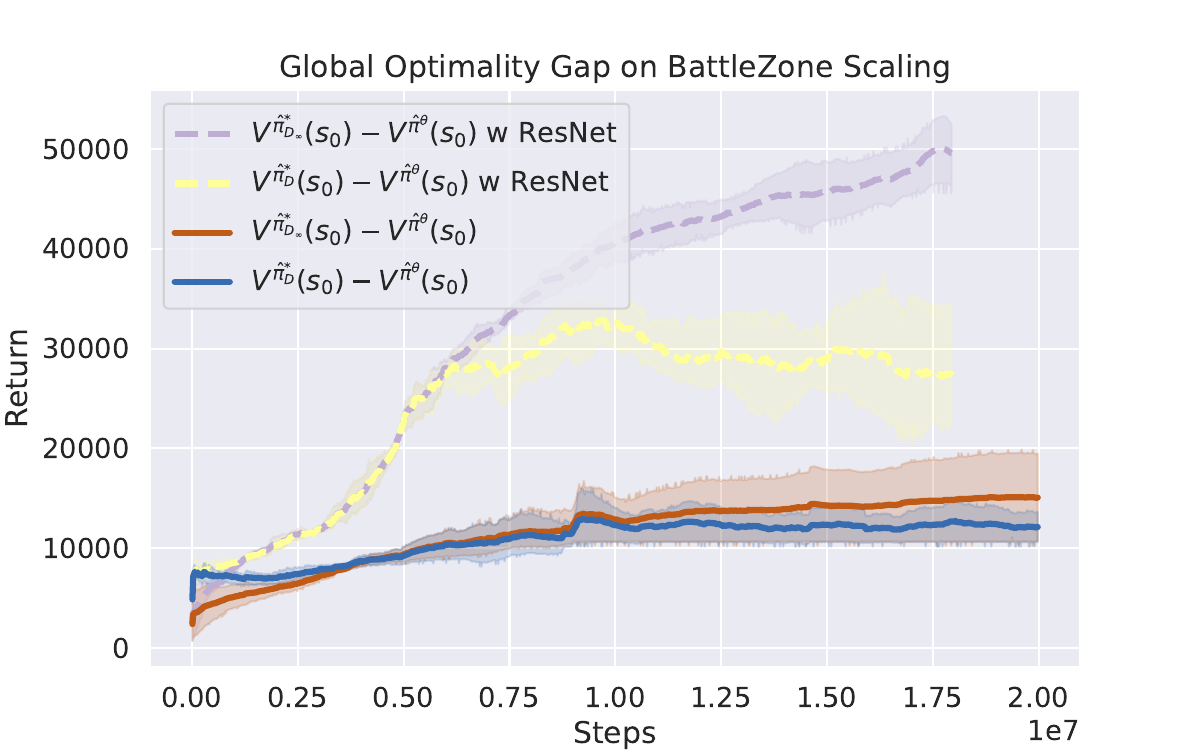}}
\subcaptionbox{\label{fig:global-optimality-explorebonus-NTG} DQN NameThisGame}{ \includegraphics[trim={0.0cm 0.0cm 0.0cm 0.0cm},clip,width=0.32\linewidth]{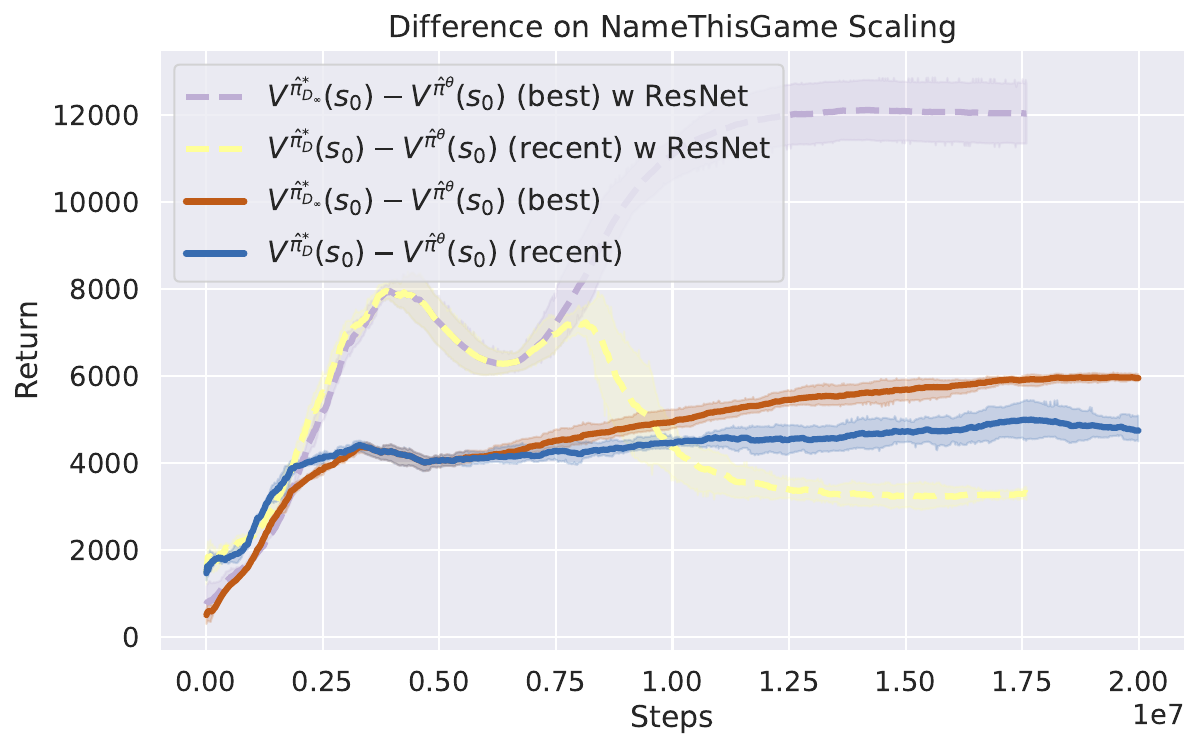}}
\subcaptionbox{\label{fig:global-optimality-scaling-cheeta} PPO HalfCheetah}{ \includegraphics[trim={0.0cm 0.0cm 0.0cm 0.0cm},clip,width=0.32\linewidth]{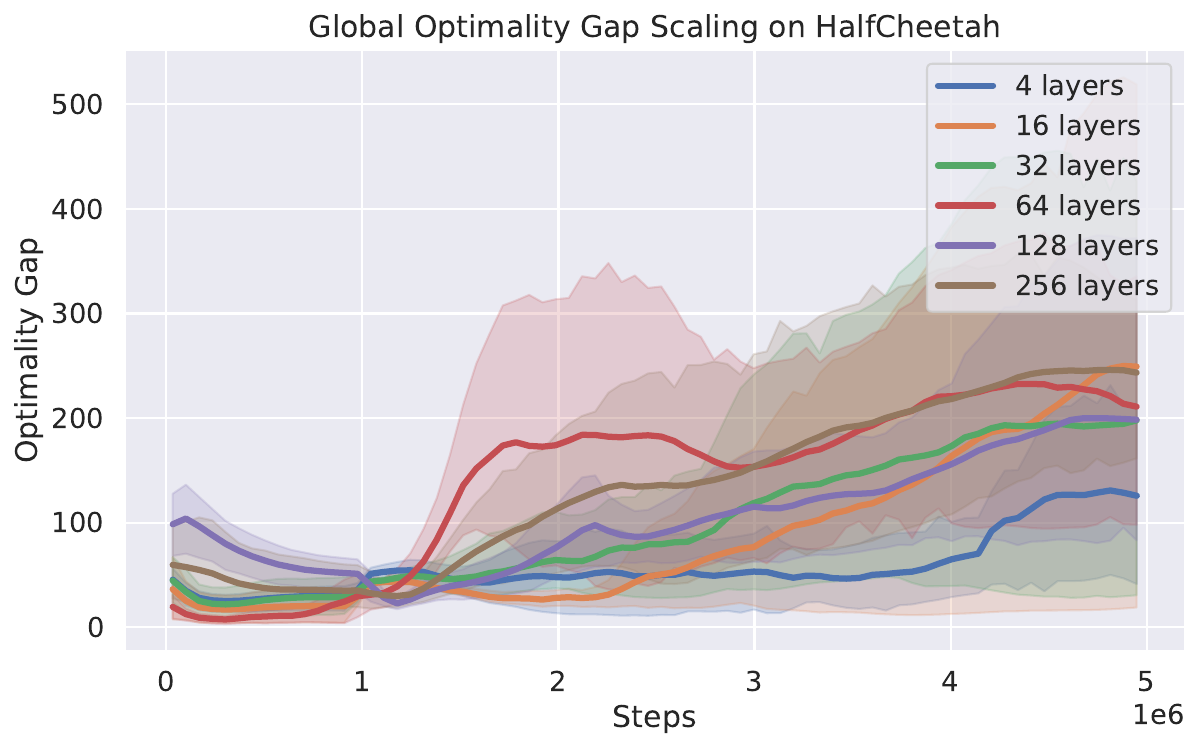}}
\caption{
Comparisons of \methodName for models with different-sized networks. On the left and middle, it is shown that using a ResNet-18 instead of the common 3-layer CNN for BattleZone increased the difference. On the right, the difference for \textbf{HalfCheetah} increases as the number of layers is added, indicating increasing exploitation issues.
}
\label{fig:global-optimality-scaling}
\vspace{-0.25cm}
\end{figure*}

In \Cref{fig:global-optimality-scaling}, the results of the described experiments are given. Interestingly, the results for the Atari environments show that the difference is much larger when the policy network is a ResNet-18 instead of a 3-layer CNN. This indicates two items: one, the policy is generating higher value trajectories, but it is not adequately learning from them, and two, the gap for \topBestFive and \topBestLocal is very close, indicating that the policy is struggling to match these higher value experiences even when they are in the current replay buffer. With \textbf{HalfCheetah}, the issue of scale is studied by training a policy over networks of $6$ different sizes. In \Cref{fig:global-optimality-scaling-cheeta}, the \topBestFive $-$ \policyAverage is shown, and there is a trend that as the number of layers increases, the \methodName increases. This is interesting because in \Cref{fig:halfCheetah} the performance with one layer is given and there is no gap. The introduction of additional layers quickly introduces exploitation issues, keeping the policy from learning the same performance in \Cref{fig:halfCheetah}. This collective information suggests that scaling networks does not likely cause exploration issues, but rather reinforces the commonly understood cause of exploitation (optimization/model) issues with scale.

\subsection{Algorithm Sub-optimality}
\label{seq:results-per-alg}
\begin{figure*}[htb]
\centering
\subcaptionbox{\label{fig:global-optimality-task} \topBestFive $-$ \policyAverage across Atari-5 environments.}{ \includegraphics[trim={0.0cm 0.0cm 0.0cm 0.0cm},clip,width=0.91\linewidth]{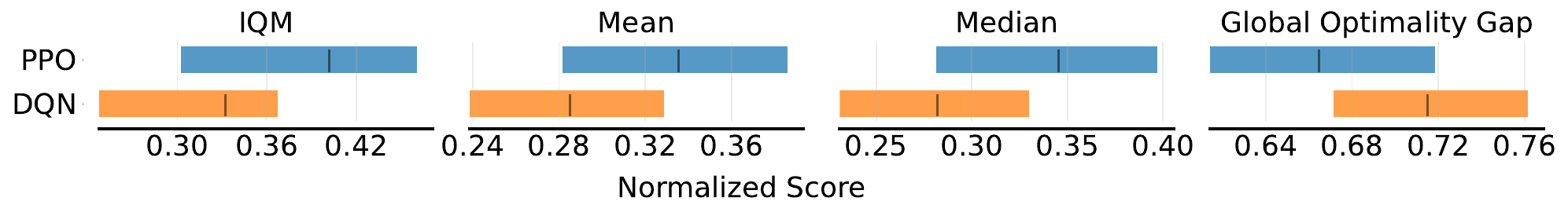}} \\
\subcaptionbox{\label{fig:aggregated-local}  \topBestLocal $-$ \policyAverage across Atari-5 environments.}{ \includegraphics[trim={0.0cm 0.0cm 0.0cm 0.0cm},clip,width=0.91\linewidth]{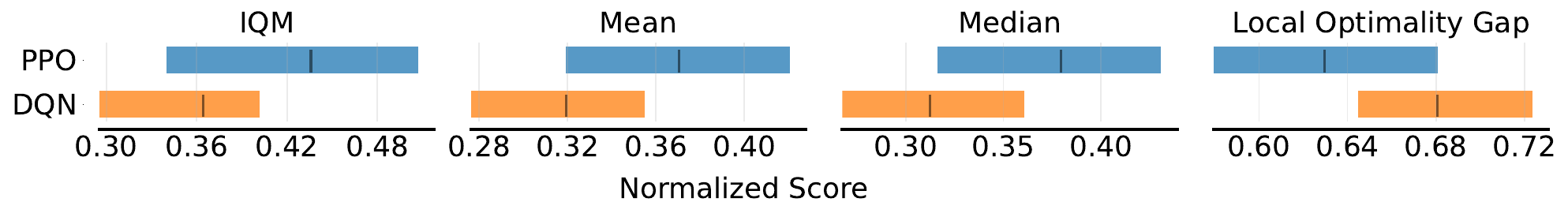}}
\subcaptionbox{\label{fig:rilable-on-atari5} Normal \textit{rliable} evaluation across Atari-5 environment.}{ \includegraphics[trim={0.0cm 0.0cm 0.0cm 0.0cm},clip,width=0.91\linewidth]{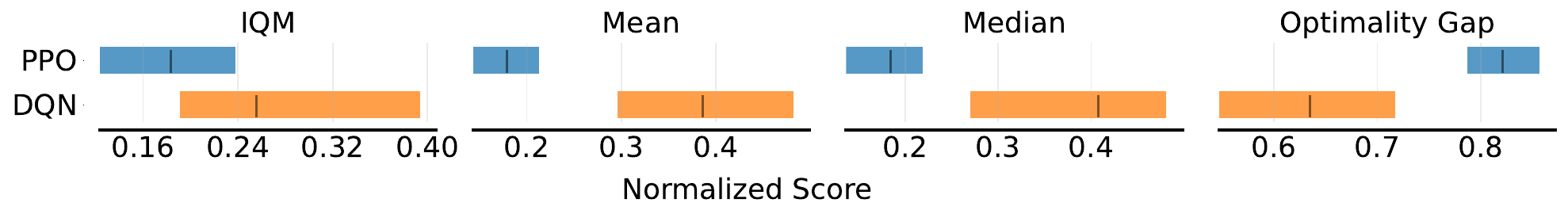}} \\
\caption{
\textit{rliable} plots for \textbf{PPO} and \textbf{DQN} over all environments in this paper. This measure gives an aggregate view for each algorithm as each sample is normalized using the \methodName from each run's generated data.
}
\label{fig:global-optimality-alg}
\vspace{-0.25cm}
\end{figure*}

Is algorithm progress limited by weaknesses in exploration or exploitation? This question can be estimated by using the \methodName to compare aggregate analysis across tasks and RL algorithms, as described in~\Cref{sec:aggregate-alg}. Starting with aggregate analysis across the \textit{AtariFive} environments, we can see in \Cref{fig:global-optimality-task} that DQN and PPO are only able to achieve a little over $30\%$ of the performance of their best experience (lower is better). This high value indicates that both of these algorithms struggle to produce the best possible results they have experienced. In this case,  \topBestFive (\Cref{fig:global-optimality-task}) is similar to \topBestLocal (\Cref{fig:aggregated-local}), indicating that the RL algorithms are experiencing high returns regularly, with a value of $0.68$, they are not sufficiently capturing. 

Interestingly and conversely, the \textit{rliable} optimality gap indicates that DQN is better than PPO in~\Cref{fig:rilable-on-atari5}, because DQN does achieve higher average policy performance, but the analysis from comparing to \topBestLocal, in \Cref{fig:aggregated-local} shows us that even though DQN performs better than PPO, DQN is still generating a lot of high-value experience that it is not able to exploit. Conversely, because PPO is performing worse according to \textit{rliable}, but has a better \topBestLocal $-$ \policyAverage, improved exploration would improve PPO more than it would DQN.
Overall, these results suggest that both algorithms struggle to extract the most from their experience and that \textit{rliable} is not telling the full story. 

\section{Discussion}
\label{sec:discussion}

This work has introduced a method to study the limitations of deep RL algorithms in the space of exploration and optimization challenges. An estimator is introduced to support this position. The estimator is used to show that common RL algorithms struggle to exploit their experience and that adding exploration bonuses and scaling networks exacerbates these issues. This estimator can be used to assist users in understanding if poor performance in an environment is the result of limited exploration (data problem) or more stable optimization to make progress (model problem).
Because RL agents collect vastly different data during training, it can be difficult to compare performance across algorithms. 
This estimator adjusts the comparison to show how well the algorithm did compared to the distribution of collected data (experience).
Because the estimator comparison over generated experience measures the sub-optimality relative to the agent's generated experience, it can be better suited to task-independent comparisons. In the future, this metric can be used to evaluate broadly across produced algorithms to assist researchers and practitioners in their analysis.

\newpage

\textbf{Acknowledgements}

I want to acknowledge funding support from Natural Sciences and Engineering Research Council (NSERC) of Canada, Samsung AI Lab, Google Research, Fonds de recherche du Québec (FRQNT) and The Canadian Institute for Advanced Research (CIFAR) and compute support from Digital Research Alliance of Canada, Mila IDT, and NVidia. I know this paper is single-authored, but that does not represent the reality of the size and number of teams that contributed to producing this research. The Mila IDT team is truly awesome and made it very easy for me to utilize a significant amount of GPU time. My lab and students for broadening the scope of these ideas and their drive to have a positive impact on this world. Authors of the cleanrl library. My assistant, Anjelica, and other administrative staff at UdeM, Mila, and CIFAR, who help make it possible for me to have time for such endeavours. The RL-sofa group at Mila, which keeps the RL discussion going. The RLC community that reminds me that RL is awesome, giving me energy to feed from. My Ph.D. supervisor, Michiel van de Panne, to whom I owe most of my inspiration and research attitude. I am sure I am missing many more.

\bibliographystyle{plainnat}
\bibliography{references} 
\clearpage

\newpage

\end{document}